%% file: main.tex
\pdfoutput=1

\documentclass[11pt]{article}

\usepackage[final]{acl}

\usepackage{times}
\usepackage{latexsym}

\usepackage[T1]{fontenc}

\usepackage[utf8]{inputenc}

\usepackage{microtype}

\usepackage{inconsolata}

\usepackage{graphicx}

\usepackage{hyperref}

\title{ERVQA: A Dataset to Benchmark the Readiness of Large Vision Language Models in Hospital Environments}

\author{
  \textbf{Sourjyadip Ray\textsuperscript{1}},
  \textbf{Kushal Gupta\textsuperscript{2}},
  \textbf{Soumi Kundu\textsuperscript{3}},
  \textbf{Payal Arvind Kasat\textsuperscript{4}},
\\
  \textbf{Somak Aditya\textsuperscript{1*}},
  \textbf{Pawan Goyal\textsuperscript{1*}}
\\
  \texttt{\{sourjyadipray@kgpian, payal@bcrmrc, saditya@cse, pawang@cse\}.iitkgp.ac.in}
  \\ 
  \texttt{kushal.oncology@aiimskalyani.edu.in, soumi.paediatrics@aiimsdeoghar.edu.in}
 \\
  \textsuperscript{1}Indian Institute of Technology, Kharagpur \\
  \textsuperscript{2}All India Institute of Medical Sciences, Kalyani \\
  \textsuperscript{3}All India Institute of Medical Sciences, Deoghar\\
  \textsuperscript{4}Dr. B.C. Roy Multi-Speciality Medical Research Centre, IIT Kharagpur \\
  }

\begin{document}
\maketitle
\begin{abstract}
The global shortage of healthcare workers has demanded the development of smart healthcare assistants, which can help monitor and alert healthcare workers when necessary. We examine the healthcare knowledge of existing Large Vision Language Models (LVLMs) via the Visual Question Answering (VQA) task in hospital settings through expert annotated open-ended questions. We introduce the Emergency Room Visual Question Answering (ERVQA) dataset, consisting of <image, question, answer> triplets covering diverse emergency room scenarios, a seminal benchmark for LVLMs. By developing a detailed error taxonomy and analyzing answer trends, we reveal the nuanced nature of the task. We benchmark state-of-the-art open-source and closed LVLMs using traditional and adapted VQA metrics: Entailment Score and CLIPScore Confidence. Analyzing errors across models, we infer trends based on properties like decoder type, model size, and in-context examples. Our findings suggest the ERVQA dataset presents a highly complex task, highlighting the need for specialized, domain-specific solutions.
\end{abstract}

\input{latex/Chapters/intro}
\input{latex/Chapters/relatedwork} 
\input{latex/Chapters/dataset} 
\input{latex/Chapters/errors}

\input{latex/Chapters/methods} 
\input{latex/Chapters/results}
\input{latex/Chapters/conclusion}
\input{latex/Chapters/ethics}

\bibliography{custom}

\appendix

\input{latex/Chapters/appendix}

\end{document}

%% file: latex/Chapters/intro.tex
\section{Introduction}

\begin{quote}
“Across globe, 6.4 million physicians needed in 132 countries facing shortages”  \emph{\href{https://www.medicaleconomics.com/view/across-globe-6-4-million-physicians-needed-in-132-countries-facing-shortages}{- Medical Economics, 2022} }
\end{quote}

\begin{figure}[!t]
\centering
  \includegraphics[width=0.8\columnwidth]{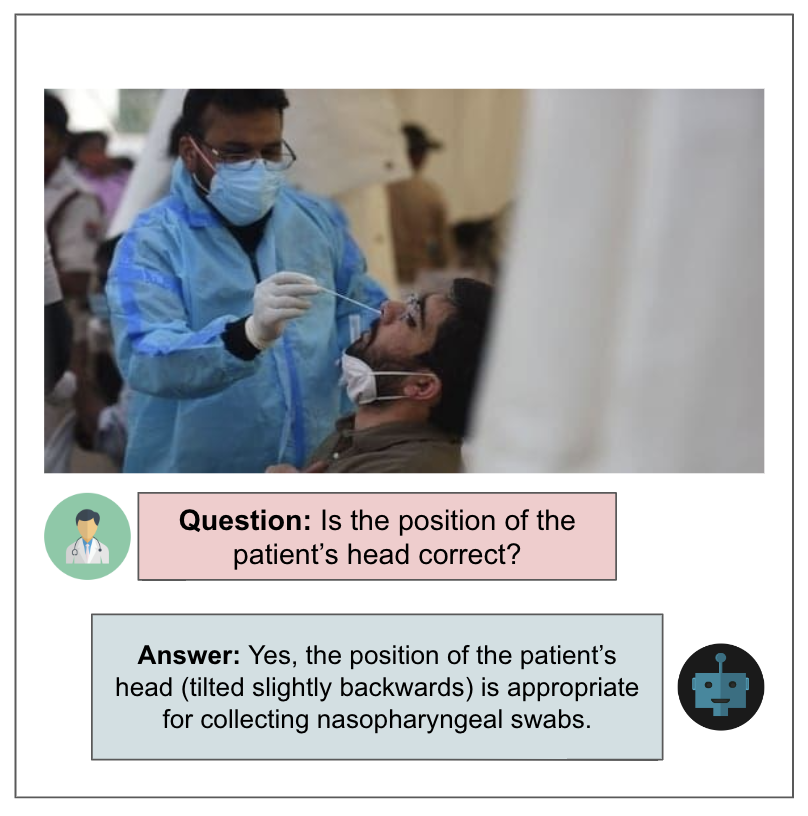}
  \caption{Example data point from the ERVQA dataset containing a manually annotated question and answer. The questions are asked from the point of view of a doctor.}
  \label{fig:intro}
  \vspace{-0.5mm}  
\end{figure}

\let\thefootnote\relax\footnotetext{* indicates equal supervision}
\let\thefootnote\relax\footnotetext{Code and data: \url{https://github.com/sourjyadip/ervqa-data/}}

The global shortage of healthcare personnel is a well-known issue \cite{world2016global}, especially severe in densely populated areas, and highlighted by crises like the COVID-19 pandemic. With the rapid growth of autonomous and remote medical agents \cite{gyles2019robots} and the increasing popularity of Vision Language models \cite{zhang2024vision}, developing smart healthcare assistants is essential. Our work explores the question: \textit{`Are the existing Large Vision Language Models (LVLMs) ready to be used in the healthcare environments?'}, focusing on their use for automated monitoring and alerting, which requires sufficient knowledge of healthcare protocols and medical diagnostic processes. 

A significant challenge in this research area is the lack of publicly available datasets. While many Visual Question Answering (VQA) datasets exist in radiology and pathology \cite{lin2023medical}, real-world datasets featuring patient and hospital environment images are scarce due to data acquisition and publication difficulties. To address this, we introduce the Emergency Room Visual Question Answering (ERVQA) dataset, featuring 4355 expert-annotated question-answer pairs based on visually plausible scenarios in emergency rooms and wards. We select images showing various patient-related scenarios, and our expert annotators formulate questions that a medical expert would typically consider and act upon. We show an example in Figure \ref{fig:intro}. 

Using our benchmark, we investigate how current state-of-the-art LVLMs (both general purpose and medical domain specific) answer the questions posed in the dataset. We observe that answers to such open-ended questions in a hospital setting should not only be relevant, but also exhibit appropriate clarity and measured caution.
A failure to meet this criterion often leads to erroneous generations.
Hence, to formally assess the performance of the generative models, we also establish a detailed error taxonomy based on common patterns, and  manually annotate generated answers with error labels. 
We study the co-occurrence patterns of these errors and gain useful insights into the non-trivialities of this problem such as the tendencies of the models to double down on erroneous generations, and the close relationships between various errors (such as Perception and Hallucination) leading to their co-occurrences. 

Keeping with the complexities associated with answering such questions and the VQA task in general, we propose two adapted evaluation metrics: \textit{Entailment Score} and \textit{CLIPScore Confidence}, useful for comparing ground truth and generated answers in this problem domain.  
In order to understand the inherent capabilities and tendencies of such models in these settings, we perform a benchmark evaluation of the dataset using various open-sourced as well as closed models, in both zero shot, as well as few-shot/in-context settings. Further analyzing error occurrences across models using finetuned BLIP-2 model-based silver labeling, we try to infer performance trends across inherent properties of LVLMs, such as decoder type, model size and in context examples. 

To summarize, our major contributions are: 1) presenting the ERVQA dataset for Visual Question Answering in healthcare, focused on emergency rooms and patient wards; 2) establishing an error taxonomy for healthcare question answering and conducting a human-annotated study on state-of-the-art models; 3) benchmarking the ERVQA dataset on various state-of-the-art LVLMs using standard and proposed metrics, as well as error analysis, providing insights for future research.

%% file: latex/Chapters/relatedwork.tex
\section{Related Work}
\begin{figure*}
    \centering
    \includegraphics[width=0.9\textwidth]{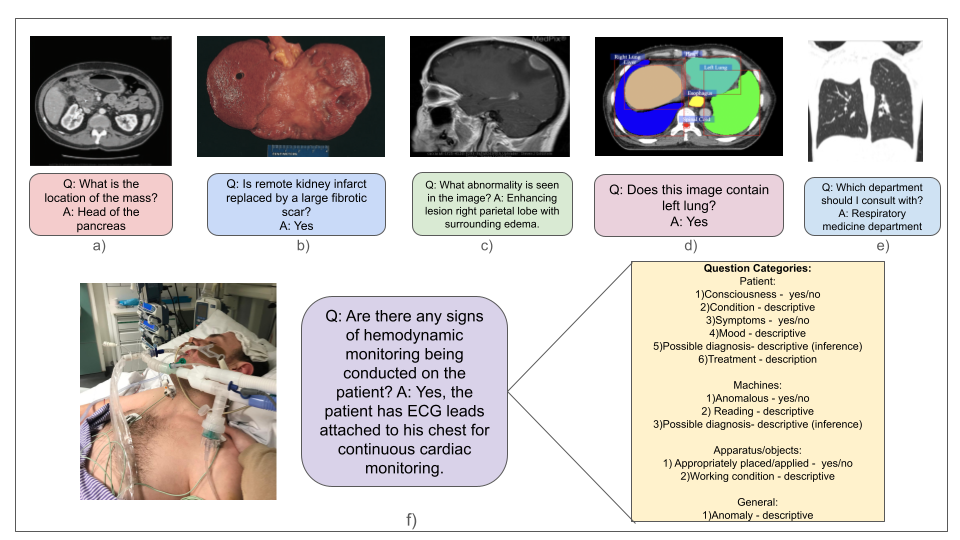}
    \caption{a) VQA-RAD dataset \cite{lau2018dataset} b) PathVQA \cite{he2020pathvqa} c) Med VQA dataset \cite{ben2021overview} d) SLAKE dataset \cite{liu2021slake} e) Patient oriented dataset \cite{huang2023medical} f) ERVQA dataset along with question categories. Annotators were asked not to limit themselves to just these categories. [\textit{ Disclaimer: All dataset images and QA pairs are taken from their respective papers.}]}
    \label{fig:comp}
\end{figure*}

\subsection{Visual Question Answering Datasets}
The Visual Question Answering (VQA) dataset \cite{antol2015vqa} presented the task for the very first time, in the multiple choice and open ended formats (open ended in this case means the answer is amongst a set of fixed tokens appearing in the training set). Subsequently, \newcite{goyal2017making} introduced VQA 2.0, which is a balanced (unbiased) version of the task involving complementary questions. Since then, various challenging datasets have been introduced by involving reasoning \cite{johnson2017clevr, hudson2019gqa, zhang2019raven} and incorporating external knowledge \cite{wang2017fvqa}. Open knowledge question answering has also been explored by \newcite{schwenk2022okvqa}. However, research on free-form VQA with open-ended long-form answers has been limited \cite{dua2021beyond} -- with some explorations towards generating answers and explanations together (VQA-E). 
\subsection{Medical Visual Question Answering}
Medical VQA has been extensively explored in radiology, microscopy, and pathology. The VQA-RAD dataset \cite{lau2018dataset} focuses on radiology images, mostly with categorical answers of 5-7 words. In contrast, we aim for longer, free-form answers beneficial to medical personnel. The PathVQA dataset \cite{he2020pathvqa} has open-ended answers averaging 2.5 words, based on questions from medical textbooks. The VQA Med dataset \cite{ben2021overview} focuses on radiology abnormalities, while the SLAKE dataset \cite{liu2021slake} uses radiology images from various body parts. Patient-oriented VQA has been explored using radiology and pathology images with template-generated questions and classification-based answers \cite{huang2023medical}. Our ERVQA dataset takes a real-world healthcare approach with emergency room images. It differs from traditional Medical VQA datasets by targeting diverse objectives (e.g., patient condition, machine readings, diagnoses), including various image types, and containing real-world, sometimes noisy, images. These factors make ERVQA more challenging. Figure \ref{fig:comp} illustrates the differences between our dataset and others.


%% file: latex/Chapters/dataset.tex
\section{ERVQA Dataset}

The Emergency Room Visual Question Answering (ERVQA) dataset consists of 4355 distinct <image,question,answer> triplets, containing real world images collected from hospital environments, annotated with hypothetical, visually plausible question-answer pairs. The dataset statistics is provided in Table \ref{tab:Table 1}.
We consult with three medical experts from well-reputed medical institutions for annotations, defining the metrics, and error classes, etc. We refer to them as expert annotators or domain experts from now on. The data collection, annotation and dataset statistics are as follows.

\begin{table}[htbp]
\centering
\small \begin{tabular}{|l|l|}
\hline
Number of QA pairs  &4355  \\ \hline
Number of images &367 \\ \hline
Average number of questions per image &11.86 \\ \hline
Average number of words per question &12.01 \\ \hline
Average number of words per answer &21.72 \\
 \hline
\end{tabular}
\caption{Dataset Statistics of ERVQA }
\label{tab:Table 1}
\end{table}

\vspace{-1em}
\subsection{Image Curation}
We selected 20 search phrases to scrape images from Google Images \footnote{\url{https://images.google.com/}}, such as `injury in hospital,' `paralysis patient in hospital,' and `stroke patient in hospital'. We scraped the top 50 results for each query and filtered images based on these criteria: 1) publicly available sources, 2) featuring a patient, healthcare worker, or medical apparatus, 3) no watermarks, and 4) real-world scenarios (not stock photos). After removing duplicates, we had 266 images for annotation.

\subsection{Question Answer Annotation}
The annotation of question answer pairs has been done using two methods: manual annotation and semi-automatic annotation. The annotators for both methods were selected based on the following set of criteria: 1) at least undergraduate level formal medical education, 2) experience in working in real world hospital settings, 3) at least bilingual proficiency in English language. All our domain experts adhere to these requirements. \\
\textbf{Manual Annotation:} Each annotator received a subset of the curated images and the following guidelines: 1) Ask questions requiring medical knowledge and expertise. 2) Formulate questions answerable using visual cues. 3) Include specific features from the image in the question. 4) Derive questions from visual inferences. 5) Avoid complex or compound questions. 6) Provide brief explanations in answers when needed. These guidelines were illustrated with visual examples. Annotators were also given common question categories and subjects for reference, but were not limited to them. \\
\textbf{Semi-Automatic Annotation:} We generate QA pairs using GPT-4V by providing the same annotation instructions as the manual annotation, via prompting, generating five QA pairs per image. These QA pairs are then manually filtered by domain specialist annotators using a two step process: 1) Verification of question validity and adherence to annotation guidelines. 2) Manual rectification of answers to valid questions. 
\begin{table}[htbp]
\small \begin{tabular}{|lllll|}
\hline
 \textbf{Metric}  &\textbf{A1}  &\textbf{A2}  &\textbf{Agree} &\textbf{Disagree}  \\ \hline
 \textit{Relevance} &93.50  &91.50  &86.00  &1.00 \\ 
\textit{Clarity}  &80.50  &86.00  &80.00  &1.50 \\ 
\textit{Harmlessness}  &81.00  &93.00  &80.00  &1.00 \\ \hline
\end{tabular}
\caption{\textit{Quality Check:} The `A1' and `A2' columns shows the percentage of `Yes' labels given individually by the two annotators. The `Agree' column shows the percentage of data points where both the annotators chose `Yes'. The `Disagree' column shows the percentage of data points where both the annotators chose `No'. }
\label{tab:Table 2}
\end{table}

\subsection{Data Augmentation}
In order to increase the linguistic diversity of the collected questions, we use GPT4V to paraphrase the questions. We collected 4 paraphrased versions of each question, which are then manually verified by the medical experts to check for the following criteria: 1) New terms used in the paraphrased questions should not change the meaning of the question 2) The questions should imply the need for the same answer as originally annotated.

\subsection{Quality Check}
As the answers to the questions of the dataset are long and sometimes require reasoning depending on the type of question, traditional inter annotator agreement metrics are not suitable. Instead, we defined the following attributes related to appropriateness of the answer, in consultation with the domain experts. \\ 
\textit{Relevance:} Whether the answer directly addresses the question in the context of the provided image. \\
\textit{Clarity:} The comprehensibility of the answer, including grammar, style, absence of ambiguity. \\
\textit{Harmlessness:} The potential for an answer to cause harm if acted upon, particularly in critical medical settings. 

We randomly sample 200 data points from the annotated data, and ask two domain experts to annotate on these metrics with a simple, categorical Yes/No label. The results are shown in Table \ref{tab:Table 2}.    
In this evaluation, for some data points, the labels given by the annotators were subjective, based on personal preference of detail, especially for the \textit{Clarity} metric. However, the extremely low percentage of `common disagreement' for all 3 metrics show that the dataset is of high quality and is reliable.   


\subsection{Dataset Statistics} 
We collect a total of 1097 question answer pairs from 367 distinct images. Out of the total number of questions, 325 QA pairs (29\%) are collected using the manual method, while the rest 772 (71\%) are collected using the semi-automatic annotation method. The dataset was then augmented by selectively paraphrasing the questions. The final number of question answer pairs is 4355. Furthermore, we find that the average number of questions per image to be 11.86. The average number of words per question is 12.01, and the average number of words per answer is 21.72.

%% file: latex/Chapters/errors.tex
\section{Error Analysis}

Answers produced by generative models for questions from the ERVQA dataset are open ended, and hence, are prone to various types of errors. These errors can be both visual or textual, leading to incorrect answers for diverse reasons. As quantifying such errors are critical for real-world scenarios, we conduct a thorough error analysis on a set of generated answers from questions from the ERVQA dataset.

\subsection{Experimental Setup}
To conduct this analysis, we randomly sample 200 questions from the ERVQA dataset. We then use 2 state of the art closed LVLMs, GPT4V-o and Gemini Pro Vision, to generate answers for these questions, with the image also given as input. Due to policy related restrictions on medical images, we obtain results for 186 images (372 generated answers), which we then use for error analysis.
The error annotation for each question was done by our expert annotators, where the definitions of the errors were given, along with the image, question and generated answer. The annotator was asked to identify all possible errors applicable to the answer. We find the Cohen's Kappa score to be 0.78, after annotation. 

\begin{figure}[!t]
\centering
  \includegraphics[width=\columnwidth]{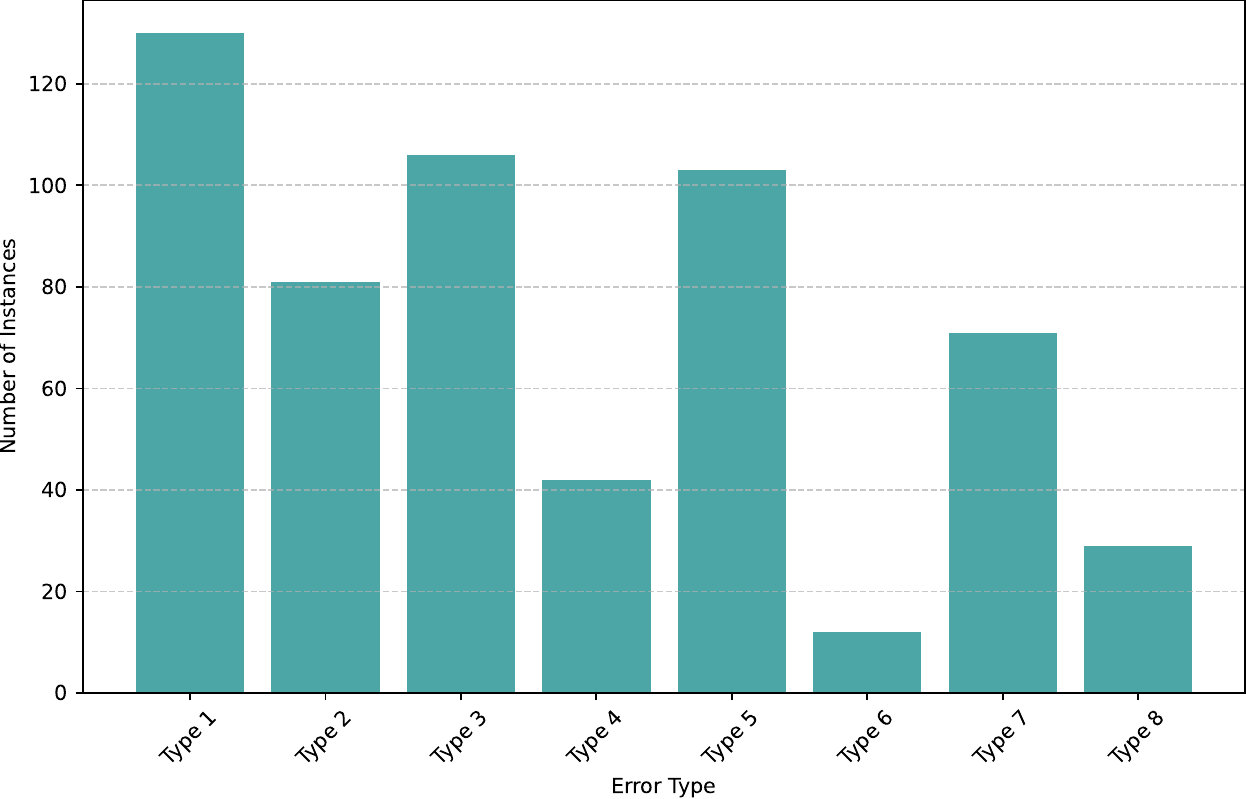}
  \caption{Distribution of error types amongst all 372 generated answers}
  \vspace{-0.5mm}
  \label{fig:errdist}
\end{figure}

\subsection{Error Taxonomy}
We define 8 error types which commonly occur in generated answers. We distinguish between these on the basis of them being fundamentally reasoning based and non-reasoning based errors. They are as follows: 

\noindent \textbf{Type 1: Reasoning Error} This error involves failure to draw logical conclusions from the image, inability to connect relevant visual cues to the question, or incorrect inference or assumption about the scene. 

\noindent \textbf{Non-Reasoning Errors: } 

\noindent \textbf{Type 2: Medical Factual Error} This error involves using incorrect medical knowledge or terminology, misunderstanding medical procedures or techniques, misinterpreting medical equipment or devices, or incorrectly interpreting patient conditions or symptoms. It stems from deficiencies in the model's healthcare and medical knowledge base.\\
\textbf{Type 3: Perception Error} This error is caused by mis-identification of objects or elements in the image, failure to recognize or distinguish medical equipment or devices, or incorrect interpretation of visual cues (e.g., patient positioning, medical staff attire). These are attributed to the model's inability to interpret visual content. \\
\textbf{Type 4: Coherence or Consistency Error} This error involves contradictory statements within the same answer. inconsistencies between the generated answer and the image, or lack of coherence or logical flow in the answer. \\
\textbf{Type 5: Specificity and Relevance Error} This error is characterized by overly general or vague answers, answering irrelevant or unrelated aspects of the question, or failure to address the specific details or context of the question. \\
\textbf{Type 6: Linguistic Error} This error involves grammatical or syntactical mistakes, inappropriate or incorrect use of medical terminology, or unclear/confusing phrasing or sentence structure. \\
\textbf{Type 7: Hallucination Error} This error is caused by generating information or details not present in the image, introducing fabricated or imagined elements in the answer, or providing responses that are entirely unrelated to the image or question. This error is fundamentally linked to the model's generation process.\\
\textbf{Type 8: Uncertainty and Confidence Error} This error is characterized by failure to acknowledge the limitations or uncertainty of the generated answer, overconfidence or lack of appropriate uncertainty estimation, or inability to distinguish between confident and uncertain responses.

\subsection{Error Occurrence and Co-occurrence} 
After collecting the error annotations for all 372 generated answers, from Figure \ref{fig:errdist} we find that \textit{Type 1}, \textit{Type 3} and \textit{Type 5} occur the most, each occurring more than 100 times (>~26\% of the generations!). \textit{Type 1} errors occur due to the models making incorrect assumptions about factors like patient history, general hospital protocols, etc. \textit{Type 3} errors mostly happen due to mis-identification of medical apparatuses, using incorrect naming of devices, and poor spatial understanding of the scene. \textit{Type 5} errors occur usually due to the model not being able to understand the motive of the question, and resorting to generating unnecessarily long answers containing definitions and general protocols about the subject. \\
\begin{figure}[!t]
\centering
  \includegraphics[width=0.9\columnwidth]{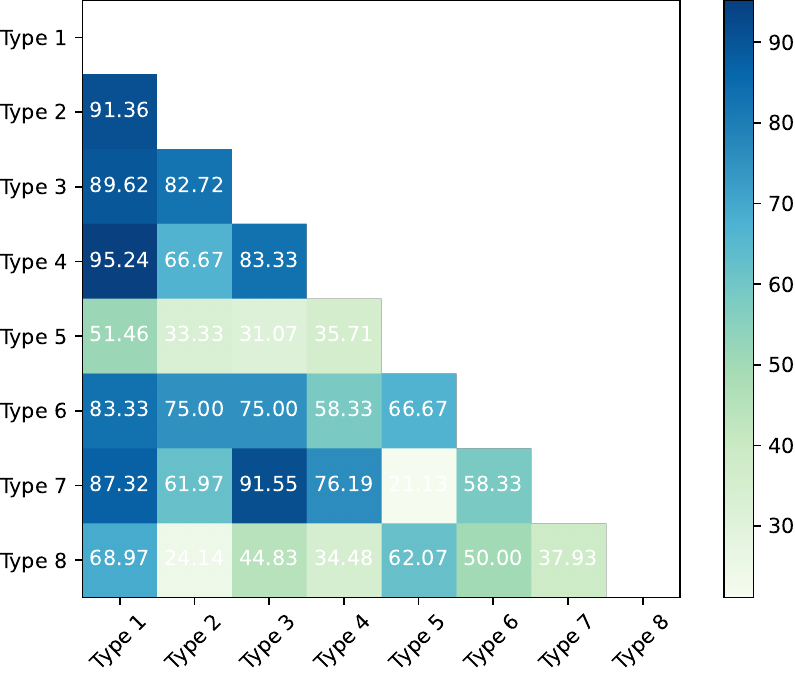}
  \caption{Error Co-occurrence Statistics}
  \label{fig:occ}
\end{figure}
\textbf{Co-occurrence Statistics:} Figure 6 shows the error type co-occurrence statistics for the error annotations. We calculate occurrence between 2 errors in the following way: For an unordered pair of errors $\{e_{i}, e_{j}\}$ where \textit{i $\neq$ j}, we count the number of co-occurrences of  $e_{i}$ and $e_{j}$, divide it by $min\{|e_{i}|,|e_{j}|\}$, and multiply by 100 for scaling, as previously shown in \newcite{roy2024beyond}. \\
\textbf{Analysis:} From Figure \ref{fig:occ}, we see that Reasoning Error (\textit{Type 1}) has a very high co-occurrence percentage with Medical Factual Errors (91.36\%), Perception Errors (89.62\%), Coherence/Consistency Errors (95.24\%), Linguistic Errors (83.33\%) and Hallucination Errors (87.32\%). This shows that GPT4Vo and Gemini Vision Pro will hallucinate, make linguistic and visual errors, and make unsure generations, and will try to \textbf{\textit{double down}} on these errors, leading to reasoning errors in the generated answer. This tendency of overconfidence, has been previously shown in the case of LLMs \cite{xiong2023can, groot2024overconfidence}. We also find from Figure \ref{fig:occ}, that Medical Factual errors (\textit{Type 2}) are also closely related to both Perception Errors (82.72\%) and Linguistic Errors (75\%), thereby motivating the fact that both linguistic and visual improvements are crucial for medically factual generation. We also see that Perception Errors (\textit{Type 3}) co-occur largely with Hallucination Errors (91.55\%), due to the two being closely related.

%% file: latex/Chapters/methods.tex
\section{Benchmark}
In order to benchmark models against the ERVQA dataset, we choose to test on models which offer a wide range of variation in the vision encoder, LLM decoder, model size and finetuning strategies. We aim to explore the inherent capabilities of these models to provide error-free meaningful answers which are lexically and semantically close to the ground truth answers, void of the errors defined in Section 4, which is essential for their application in the healthcare domain. 

\subsection{Models}
We experiment with both open sourced and closed models. Open source models include Llava 1.5 13B \cite{liu2024improved} which consists of a CLIP ViT-L/14 encoder and Vicuna-7B decoder, Open-Flamingo 9B \cite{awadalla2023openflamingo} consisting of CLIP ViT-L/14 encoder and Llama-7B decoder, MPLUG-Owl 7B \cite{ye2023mplug} consisting of CLIP ViT-L/14 encoder and MPT-7B decoder, and Instruct-BLIP 7B model \cite{dai2024instructblip} containing a BLIP-2 based encoder and Vicuna-7B decoder. We experiment with Med-Flamingo 9B model \cite{moor2023med} with Llama-7B decoder, finetuned on medical VQA datasets, to compare with its open domain counterpart Open-Flamingo. Although the medical data used is from the pathology/radiology domain (See Section 2), our interest lies in finding out whether it captures the nuances of answering medical questions. Finally, we also run experiments on closed models GPT4Vo by OpenAI \cite{achiam2023gpt} and Gemini Pro Vision \cite{reid2024gemini} by Google, to understand the capabilities of larger state of the art models trained on more training data.    

\subsection{Experimental Setup}
Our goal is to understand the \textit{inherent} capabilities of LVLMs with our challenging dataset. We run experiments in zero-shot and in-context few-shot (1-shot and 3-shot) settings to assess these models in low-resource scenarios. Llava-1.5 and InstructBLIP are tested only in zero-shot, and Open-Flamingo only in few-shot settings, due to limitations in their training strategies. The requirement for in-context examples for Open-Flamingo in 'zero-shot' settings is supported by \cite{awadalla2023openflamingo}. All the prompts, code and data have been made available for reproducibility (see Appendix).

\subsection{Evaluation metrics}
\label{sec:eval_metrics}
To evaluate generated answers against ground truth, we use traditional model-free metrics such as \textbf{BLEU-1} \cite{papineni2002bleu}, which compares unigrams, and \textbf{ROUGE-L} \cite{lin2004rouge}, which compares longest common subsequences. For semantic similarity, we use \textbf{SentenceBERT Similarity} \cite{reimers2019sentence}. However, these metrics do not fully capture two essential factors for VQA evaluation: a) whether the ground truth answer is entailed within the generated answer, and b) whether the generated answer aligns with the image. To address these issues for our dataset, we adpapt two model based evaluation metrics suitable for our problem domain:\\
\textbf{Entailment Score:} We use the \textit{roberta-base-nli} model \footnote{ \url{https://huggingface.co/cross-encoder/nli-roberta-base}}, trained on the SNLI \cite{bowman2015snli}  and MultiNLI \cite{williams2017broad} datasets. We define Entailment score as follows:
\begin{equation}
 ES = p(entailment|ref, gen)
\end{equation}
We obtain ES by applying a softmax function on the entailment class logit obtained from the \textit{roberta-base-nli} model, and we report the average across all the generations. \\
\textbf{CLIPScore Confidence:} CLIPScore has been originally introduced as a reference-free metric for evaluating the Image Captioning task \cite{hessel2021clipscore}. We treat the reference and generated answers as if they were captions for the image, and calculate the CLIPScore using the rich cross-modal image and text embeddings, as described in the paper. We. then calculate the reference-based CLIPScore Confidence as follows:
\begin{equation}
 \scriptsize CLIP-C = \frac{CLIP-S(img,gen)}{CLIP-S(img,ref)+CLIP-S(img,gen)} \scriptsize
\end{equation}
This gives us an image-based comparison of the generated and reference answers.

\subsection{Error Evaluation}
\label{sec:error_eval}
Visual Question Answering using the ERVQA dataset is a complex task due to its sensitive applications, requiring the answers to be as error-averse as possible. Hence, evaluating models based on similarity metrics as well as occurrence of errors is of paramount importance. Since human annotations for all generations is expensive as well as cumbersome, we obtain silver-label error annotations using classification models for each type of error, and compare and analyse the performance based on percentage of error occurrences. \\
\textbf{Error Classification Model: } In order to build an error-wise dataset for building classification models, we first include all the manually annotated data obtained from Section 4 for each label. To balance the dataset, we introduce more erroneous examples by inducing errors to ground truth answers by prompting GPT4V with the error definition. Finally, for each class we obtain 200 positive and 200 negative samples. 

We finetune a BLIP-2 model \cite{li2023blip}\footnote{\url{https://huggingface.co/Salesforce/blip2-opt-2.7b}} for error classification, where we pool the cross-modal QFormer outputs and attach a binary classification head to it. After training for 10 epochs with a learning rate of 1e-6, we save the model with the best performance against the validation set. We finally pass the model logits through a sigmoid layer to get the class predictions. We use a 5-fold cross validation setting where, across all classes, we report average macro F1 score on the best runs, to be 97.81.  

%% file: latex/Chapters/results.tex
\section{Results and Analysis}
\begin{table}[!htbp]
\scriptsize
\setlength{\tabcolsep}{2.5pt}
\centering
\begin{tabular}{|l|lllll|} \hline 
 &\textbf{BLEU}  &\textbf{ROUGE}  &\textbf{SENT}  &\textbf{ENT}  &\textbf{CLIP-C}  \\ \hline
\textbf{Zero-shot} &  &  &  &  &  \\ 
mPLUG-Owl-7B &18.13  &22.32  &65.50  &23.20  &49.93  \\
Instruct-BLIP-7B* &16.82  &20.13  &51.75  &18.32  &48.44  \\
Med-Flamingo-9B &17.20  &19.17  &64.23  &28.19  &49.20  \\
Llava1.5-13B* &20.11  &22.78  &66.00  &19.03  &\textbf{52.31}  \\
GPT4Vo &16.63  &20.72  &69.34  &38.98  &51.02  \\
Gemini Pro Vision &19.83  &24.21  &64.36  &19.10  &49.87  \\ \hline
\textbf{1-shot} &  &  &  &  &  \\ 
mPLUG-Owl-7B &18.74  &23.01  &69.30  &24.47  &50.98  \\
Med-Flamingo-9B &20.76  &25.62  &61.12  &10.49  &48.95  \\
Open-Flamingo-9B &18.01  &\textbf{29.10}  &64.34  &11.18  &48.98  \\
GPT4Vo &22.25  &26.00  &76.79  &\textbf{40.63}  &51.31  \\
Gemini Pro Vision &23.27  &27.05  &68.58  &28.94  &50.38  \\ \hline
\textbf{3-shot} &  &  &  &  &  \\ 
mPLUG-Owl-7B &20.03  &26.67  &64.83  &20.04  &50.11  \\
Med-Flamingo-9B &20.21  &25.34  &61.23  &8.58  &49.56  \\
Open-Flamingo-9B &17.54  &28.70  &63.91  &9.07  &49.01  \\
GPT4Vo &23.66  &27.10  &\textbf{71.17}  &38.22  &51.03 \\
Gemini Pro Vision &\textbf{24.02}  &27.67  &68.52  &29.98  &51.23 \\ \hline 
\end{tabular}
\caption{Evaluation metrics based results for all models on zero-shot, 1-shot and 3-shot settings. \textbf{BLEU} denotes BLEU-1 score, \textbf{ROUGE} denotes ROUGE-L score, \textbf{SENT} denotes SentenceBERT Similarity, \textbf{ENT} denotes Entailment Score and \textbf{CLIP-C} denotes CLIPScore Confidence. Bold figures indicate best results across all settings. * denotes models which have only been used for zero-shot evaluation.}
\label{tab:eval_metrics_allmodels}
\end{table}
\subsection{Evaluation Metrics based Results}
Table \ref{tab:eval_metrics_allmodels} presents the results across all models and settings based on the evaluation metrics described in Section \S\ref{sec:eval_metrics}. In the zero-shot setting, larger model sizes generally yield better results across all metrics. Llava-1.5 is the best-performing open-source model with the highest CLIP-C score. GPT4Vo provides the most semantically similar answers with high entailment scores. Among the 7B models, mPLUG-Owl performs best.

In few-shot in-context settings, increasing the number of examples improves BLEU, ROUGE, and SENT scores, indicating better lexical pattern learning and word choice. However, open-source models show a decline in semantically relevant metrics like ENT and CLIP-C, indicating a tradeoff between lexical accuracy and semantic correctness. Closed models, on the other hand, improve in most metrics when provided with examples. Interestingly, there is no significant improvement from medical domain-specific fine-tuning, as seen with comparable results between Open-Flamingo and Med-Flamingo. This is likely due to Med-Flamingo being fine-tuned on Medical VQA datasets focused on pathology/radiology rather than the broader healthcare settings of ERVQA. 

\subsection{Error based Results}
In general, the values of the embedding cosine-similarity based and lexical overlap based metrics reported for open-domain datasets, such as in \citet{dua2021beyond}, are comparable with that of our experiments on ERVQA. However, in this section, we show that metric-based results do not reflect in the actual quality of the generated texts, with regards to errors. We obtain silver-labels for each error class using the classification model described in Section \S\ref{sec:eval_metrics} across all models and perform a comparison on the basis of percentage of error occurrence. We analyse our results based on variance in the following factors: a) LLM decoder type b) Model size and c) Number of In-context examples. \\
\begin{figure}[!t]
  \includegraphics[width=\columnwidth]{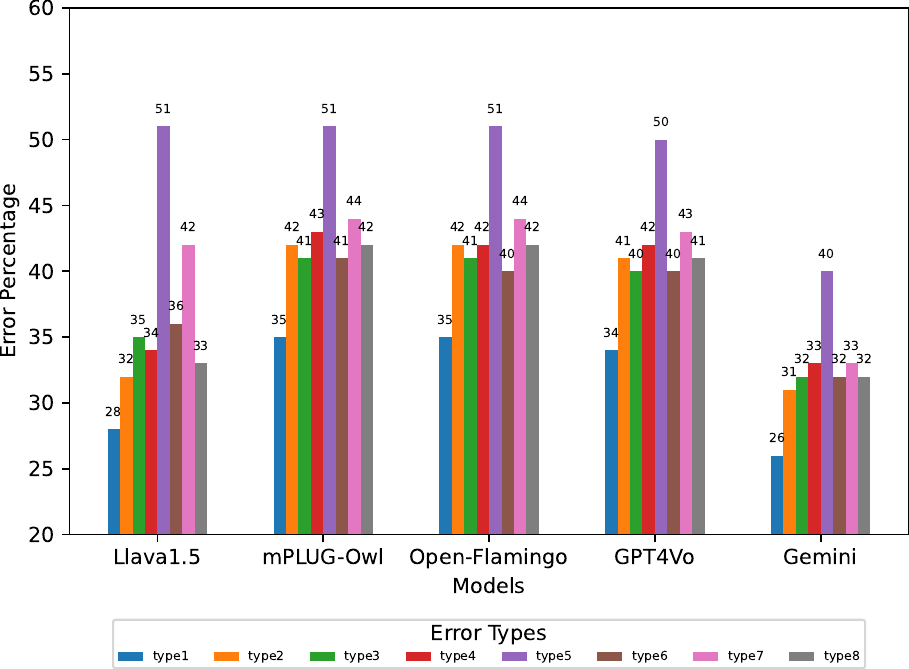}
  \caption{Model wise percentages of error occurrence for comparison on the basis of LLM Decoder type}
  \label{fig:decoder}
\end{figure}
\begin{figure}[!t]
  \includegraphics[width=\columnwidth]{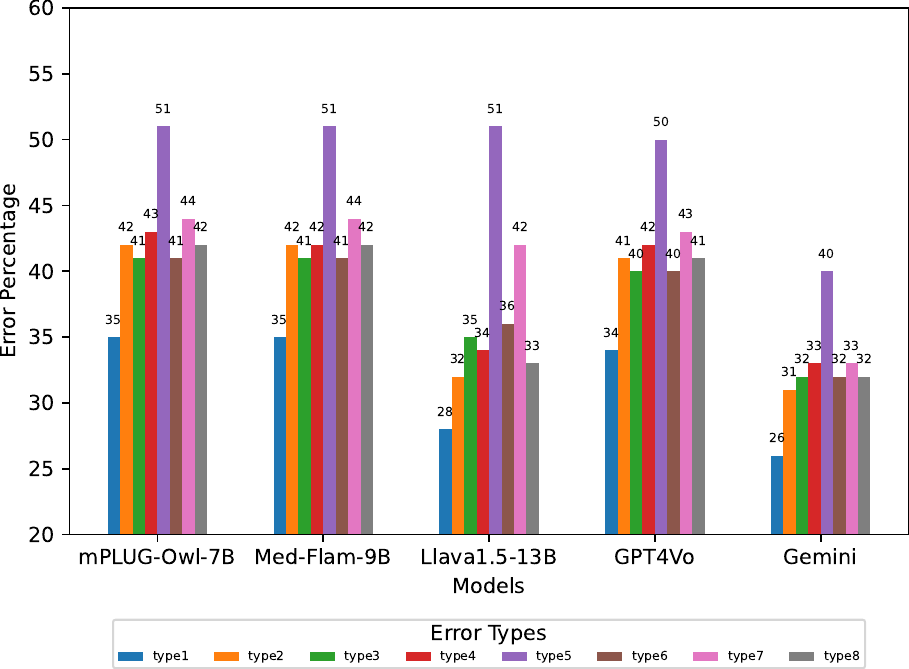}
  \caption{Model wise percentages of error occurrence for comparison on the basis of Model Size}
  \label{fig:size}
\end{figure}
\begin{figure}[!t]
  \includegraphics[width=\columnwidth]{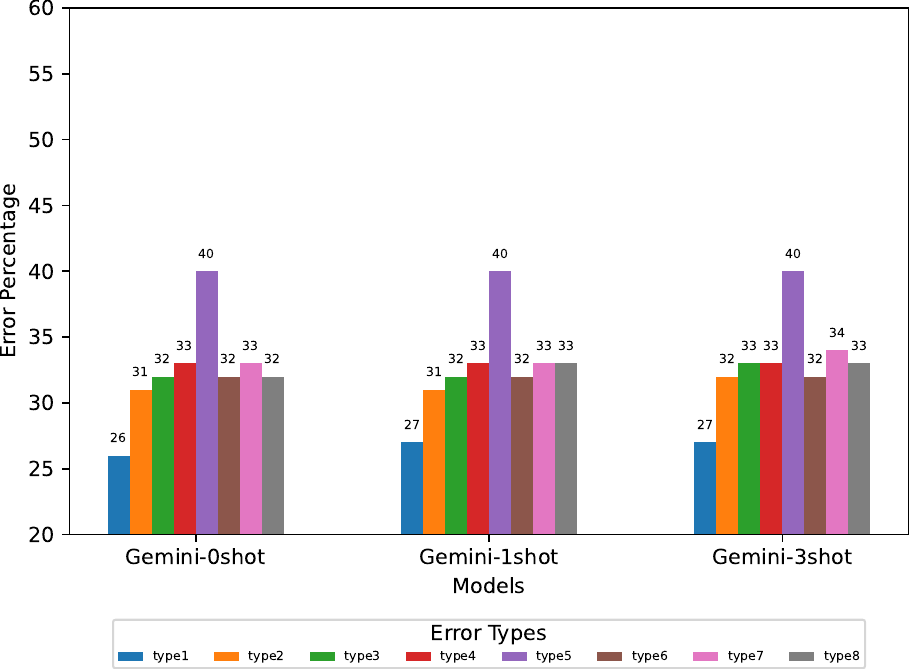}
  \caption{Model wise percentages of error occurrence for comparison on increasing in-context examples for Gemini Vision Pro}
  \label{fig:fewshot}
\end{figure}
\textbf{LLM decoder type based analysis:}
We consider 5 different best-performing LLM decoder based VLMs in the zero-shot setting for analysing the percentage of errors. Llava1.5 has a Vicuna-7B \footnote{\url{https://lmsys.org/blog/2023-03-30-vicuna/}} based decoder, mPLUG-Owl has a Llama-7B \cite{touvron2023llama} based decoder, Open-Flamingo has a MPT-7B \cite{mosaicml2023introducing} based decoder, and the closed models (GPT4vo and Gemini Pro Vision) have undisclosed decoder models \cite{achiam2023gpt} \cite{reid2024gemini}. According to Figure \ref{fig:decoder}, among open source models, we notice lesser errors in the Llava1.5 model based on Vicuna. Overall, again Gemini gives the best results. Llama-7B, MPT-7B and GPT4Vo give similar results. An important point to note is that apart from Gemini, all other models give high percentage of Specificity/Relevance error (Type 5) (>50\%), thus showing that these models have a tendency of generating long answers containing irrelevant details. In most other metrics, it is important to note that the Llava-1.5 model performs comparably to Gemini. \\
\textbf{Model size based analysis:}
We conduct experiments on different models of various sizes in the zero-shot setting. Our models, in increasing order of size are as follows: mPLUG-Owl-7B < Med-Flamingo-9B < Llava1.5-13B< GPT4Vo ~ Gemini Vision Pro. From Figure \ref{fig:size}, although we see Llava-13B having lesser errors than smaller models, GPT4Vo gives more errors. Hence, error occurrence is invariant of model size. \\
\textbf{Effect of In-context Few shot examples:}
From Table 2, we see that increasing the number of in-context examples leads to an increase in performance for the Gemini Vision Pro, except for the CLIP-C metric (there is an increase from zero-shot, but not uniform). Hence, we try to answer the question: \textit{Does increase in performance in terms of evaluation metrics mean a decrease in percentage of errors?} We compare the error percentages across zero-shot, 1-shot and 3-shot settings for the Gemini Pro Vision model, and we find that the error occurrences are almost identical -- meaning that the erroneous generations have little or no correlation to token-based and semantic metrics.  

%% file: latex/Chapters/conclusion.tex
\section{Conclusion and Future Work}
Finally, we revisit our question: \textit{"Are the existing Large Vision Language Models (LVLMs) ready to be used in healthcare environments?"} The ERVQA dataset is proposed as the first step towards addressing this. Our extensive error taxonomy and human evaluation reveal the complexity of generating accurate answers due to dense occurrence and co-occurrence patterns. Despite adapting semantically relevant VQA-specific evaluation metrics, improvement across these metrics doesn't necessarily reduce errors. The best-performing model, Gemini Vision Pro, still has an average error rate of 33\% across all error classes, which is significant given the sensitive domain. Therefore, we conclude that \textbf{existing LVLMs are not yet ready for healthcare environments}. Our dataset is a valuable resource for evaluating and improving models and techniques in this field. We hope that the future work would focus on building more error-sensitive LVLMs, developing effective error mitigation strategies, and exploring video datasets for further advancement.

%% file: latex/Chapters/ethics.tex
\section*{Limitations}
Our work, although comprehensive and valuable, has the following limitations. First, the absence of existing high quality pre-traning data due to its highly confidential nature, has limited the knowledge base of existing LVLMs, we use for benchmarking our dataset. This affects the performance of models in consideration. Second, the doctors advising and helping with the annotations are from premium medical facilities from India. Their annotations are influenced by formal medical education and real-world experience in India. While this common background, along with strict guidelines ensure uniformity in their notions of relevance and accuracy of the answers, the experts were likely following medical protocols common in India. However, they might differ from doctors from other regions of the world in some cases, due to differences in medical standards and protocols, cultural and ethical considerations, resource availability and infrastructure, legal and regulatory frameworks and public health priorities particular in India \cite{mcpherson1989international}. A wider collaborative effort by medical experts from across the world, would significantly enhance this domain of study, both through diverse data and better modeling.

\section*{Ethics Statement}
Since the annotations of the questions and answers, along with the error annotations, have been made by multiple domain experts, it is possible that the dataset reflects social, ethical and medical biases of the individuals. \\
All images scraped for the purposes of building this dataset are taken from Google Images search results, sourced from various news articles and similar publications. We do not claim any ownership of the images itself, but of the annotations done by the medical professionals. Image URLs are provided in the dataset files for downloading. To the best of the authors’ knowledge, the images have been scraped from paywall-free articles. We plan to make the image links available on request after making the dataset publicly available, to ensure safe and responsible usage. \\
Furthermore, the questions and answers framed for this dataset are on the basis of plausible scenarios according to the image content. No real world medical records/data were used to frame the question answer pairs. Apart from the visual information in these images, we have not provided any resources for identification of the individuals such as patients/doctors/nurses and also institutions such as hospitals. We shall also provide an opt-out option for those who want certain data points censored/removed. \\
We use ChatGPT as a collaborative writing and coding assistant when required. However, all novelties including metric definitions, taxonomies, prompts, etc are completely made by the authors/collaborators.

\section*{Acknowledgements}
The work was supported in part by a research grant from IIT KHARAGPUR  AI4ICPS I HUB  FOUNDATION. PI Pawan Goyal acknowledges the kind support by the grant given by I-Hub Foundation for Cobotics, IIT Delhi. PI Somak Aditya further acknowledges the support by SERB SRG/2022/000648.

%% file: latex/Chapters/appendix.tex
\section{Appendix}
\label{sec:appendix}

\subsection{Additional Related Work}
There has been a rise in popularity of automatic evaluation of LLM generated text \cite{chiang2023closer}. \cite{lin2023llm} discusses the use of prompt based evaluation of text using LLMs. \cite{liu2024towards} discusses using automatic evaluation of clinical text. LLM based evaluation of VQA has also been explored in \cite{manas2024improving}. Other unsupervised methods like DEB \cite{sai2020improving} uses BERT as an evaluation model for adversarial text, while FED \cite{mehri2020unsupervised} uses an unsupervised method for measuring dialogue quality. We evaluate error occurrences using multi-modal representations from BLIP-2, making use of both text and visual modalities.

\subsection{ERVQA Dataset Additional Details}
\begin{figure*}[!t]
\centering
  \includegraphics[width=0.9\textwidth]{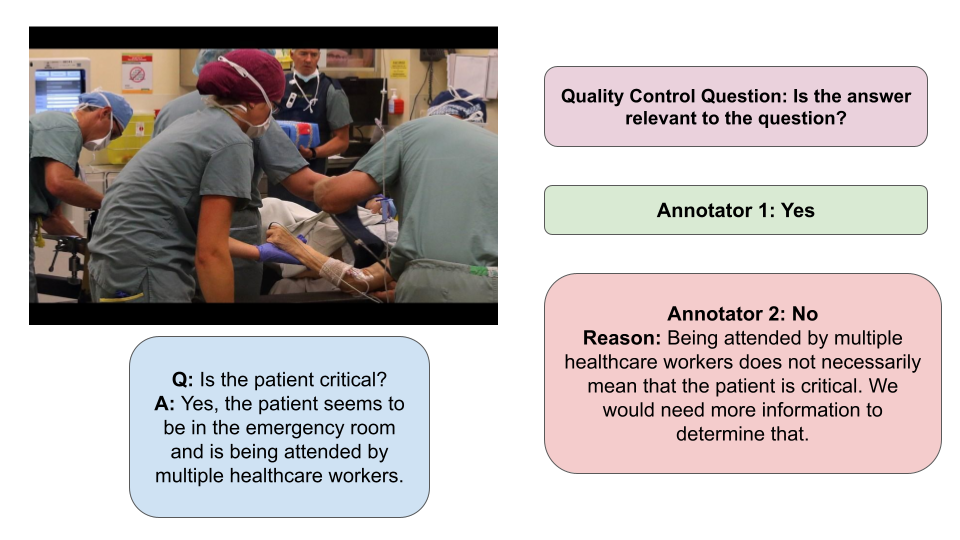}
  \caption{Disagreement on the basis of \textit{Relevance}}
  \label{fig:disagree}
  \vspace{-0.5mm}
\end{figure*}

\begin{figure}[!t]
\centering
  \includegraphics[width=0.75\columnwidth]{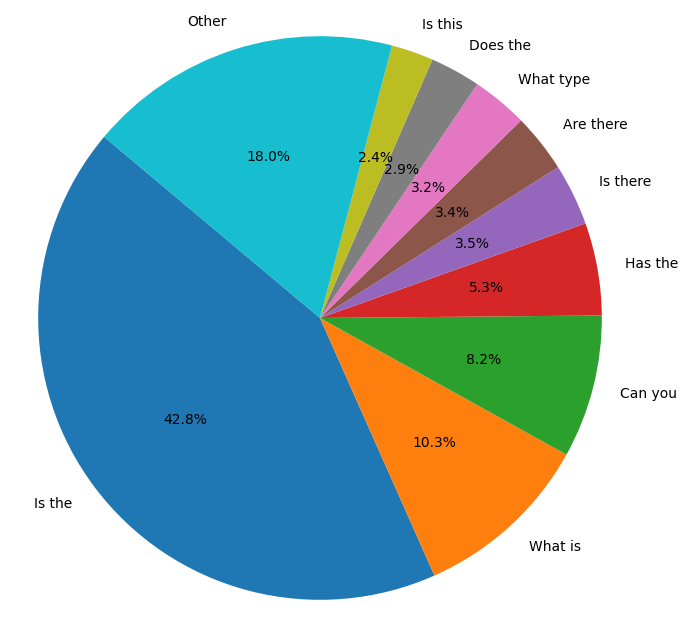}
  \caption{Distribution of the most common beginning bigrams for questions in the ERVQA dataset. The 'other' section is very diverse containing bigrams like \textit{'How frequently', 'Please report, 'Which setup'}, etc}
  \label{fig:piechart}
  \vspace{-0.5mm}
\end{figure}

During annotation with regards to the qualitative metrics: \textit{Relevance}, \textit{Clarity} and \textit{Harmlessness}, the annotators sometimes disagree with each other. One such disagreement is shown in Figure \ref{fig:disagree}. \\
The ERVQA dataset consists of a wide range of questions of varying types. A pie chart showing the various different types of questions are shown in Figure \ref{fig:piechart}.

\subsection{Examples of Different Errors}
\begin{figure*}[!t]
\centering
  \includegraphics[width=0.8\textwidth]{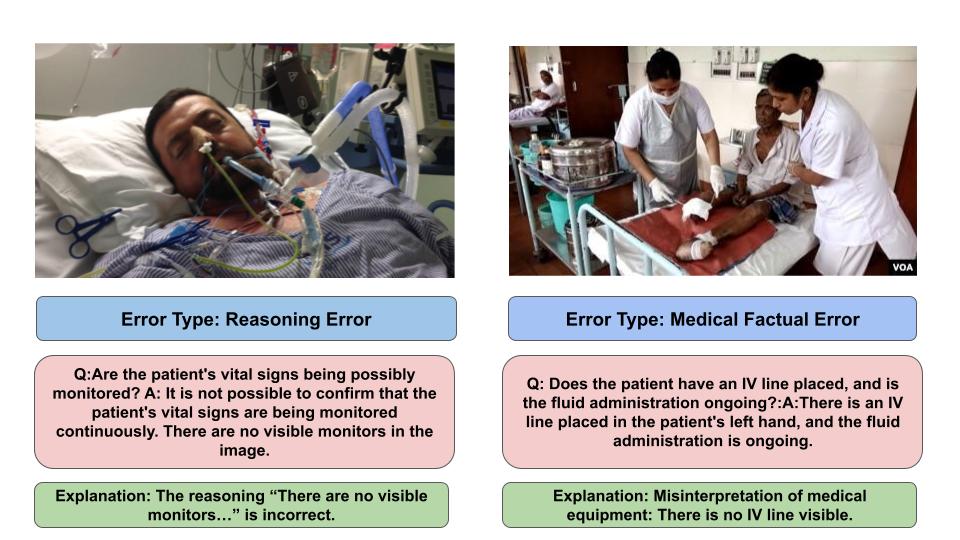}
  \caption{}
  \label{fig:err1-2}
\end{figure*}

\begin{figure*}[!t]
\centering
  \includegraphics[width=0.8\textwidth]{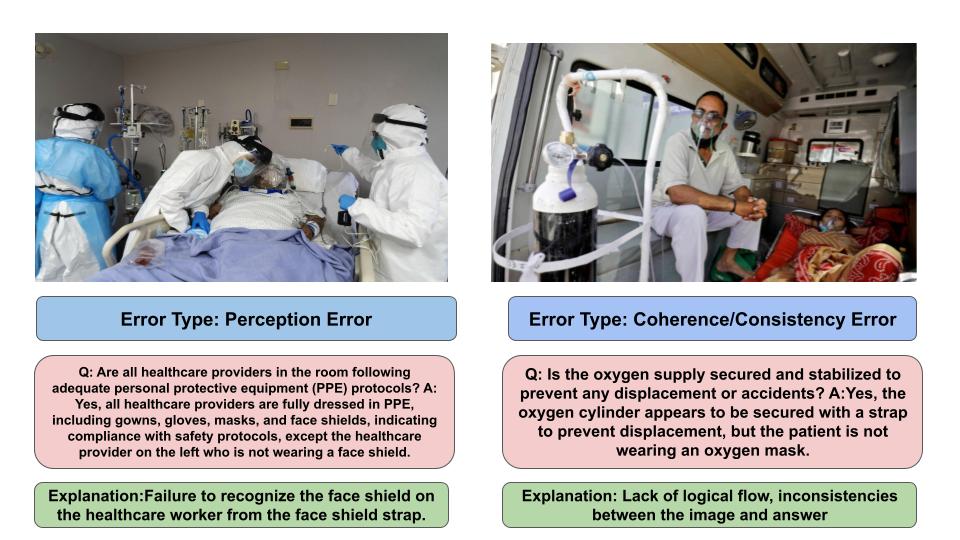}
  \caption{}
  \label{fig:err3-4}
\end{figure*}

\begin{figure*}[!t]
\centering
  \includegraphics[width=0.8\textwidth]{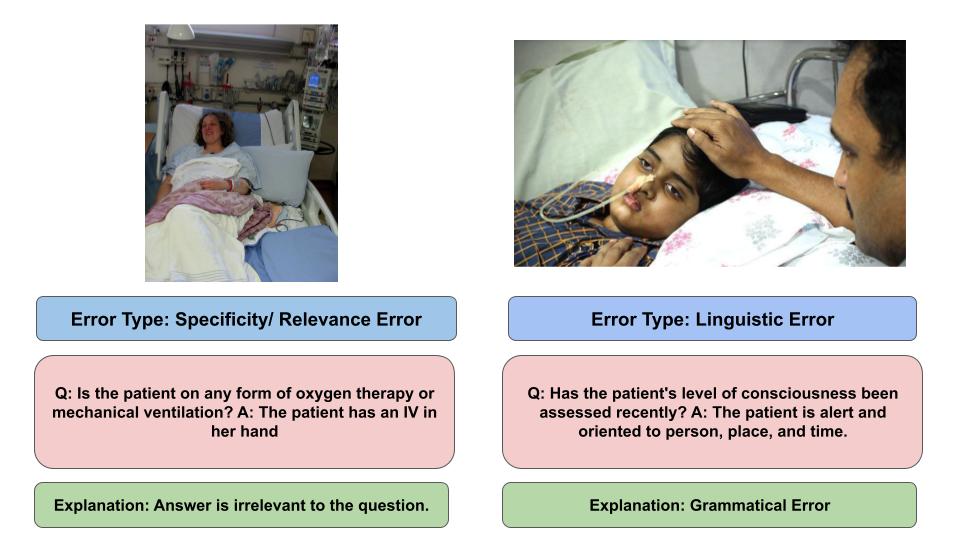}
  \caption{}
  \label{fig:err5-6}
\end{figure*}

\begin{figure*}[!t]
\centering
  \includegraphics[width=0.75\textwidth]{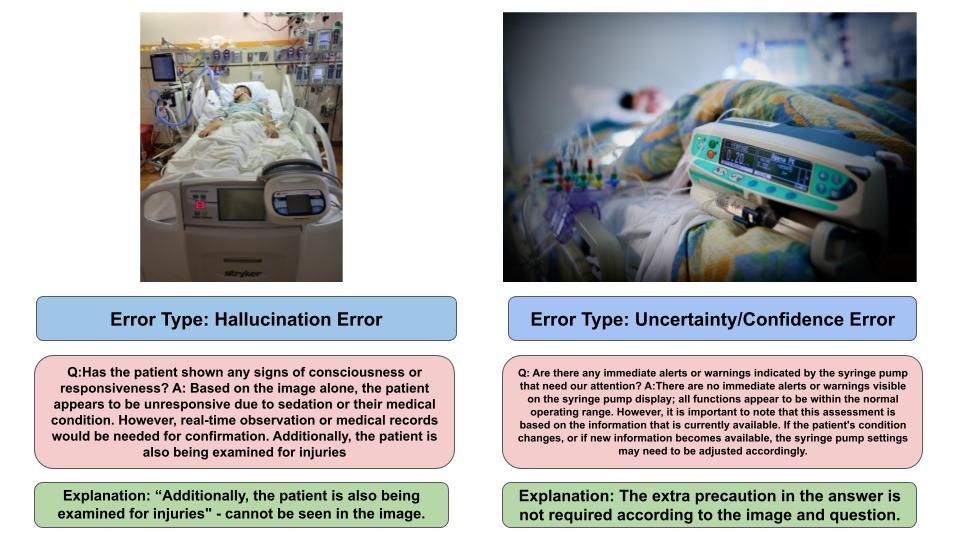}
  \caption{}
  \label{fig:err7-8}
\end{figure*}

We demonstrate the different kinds of errors from the error taxonomy with an example image, question, erroneous answer and explanation. Please note that in some cases, while a single error is pointed out, there might be other errors within the answers, that have been avoided to prevent confusion in the reasoning process. This co-occurrence of errors makes the annotation as well the prediction task extremely hard and subjective. The errors are shown in Figures \ref{fig:err1-2}, \ref{fig:err3-4}, \ref{fig:err5-6} and \ref{fig:err7-8}.

\subsection{Training Details of Error Classifier Model}
All the BLIP-2 based error classification models for the 8 types were trained on a single NVIDIA L40 GPU having 48GB GPU memory. We adopt a 5-fold cross validation strategy, and save the checkpoints with lowest Binary Cross Entropy Loss for the validation set. We fix the learning rate to 1e-6 for the Adam's optimizer after initial experimentation.  

\subsection{Reproducibility}
For reproducibility of our zero-shot and few-shot in-context results, we provide the prompts used for generation. Please note we experiment with simpler prompts at first, and increase the complexity according to documentation and failure cases. Also, we make our datasets, error classifier and model code publicly available at \url{https://github.com/sourjyadip/ervqa-data/}.
\subsubsection{GPT4Vo Prompts}
\textbf{Zero-shot: } "Answer the following question for the given image: " + <question> \\
\textbf{1-shot and 3-shot: } We use the prompting strategy used in the \href{https://community.openai.com/t/how-to-do-few-shot-prompting-interweaving-text-and-images-with-gpt-4-vision-preview-as-seen-in-the-dawn-of-lmms-preliminary-explorations-with-gpt-4v-ision/598794/3}{OpenAI Forum}.
\subsubsection{Gemini Vision Pro Prompts}
The images are provided to the model separately in batches. \\
\textbf{Zero-shot: }<question> + " Answer:"
\textbf{1-shot: }"You are being provided with 1 example question and answer for the first image. Similarly, give an answer for the second question based on the second image. Q: "+ <question1> + "A: "+ <answer1> + " Q: "+ <question2> + "A: " \\
\textbf{3-shot: }"You are being provided with 3 example questions and answers for the first 3 images. Similarly, give an answer for the fourth question based on the fourth image. Q: "+ <question1> + "A: "+ <answer1> + " Q: "+ <question2> + "A: "+ <answer2> + " Q: "+ <question3> + "A: "+ <answer3> + " Q:" + <question> + "A: "

\subsubsection{Llava Prompt}
\textbf{Zero-shot: } "<image> \textbackslash nUSER: " + <question> + "\textbackslash nASSISTANT:"

\subsubsection{Med-Flamingo Prompts}
\textbf{Zero-shot:} "You are a helpful medical assistant. You are being provided with an image and a question about the image. Answer the question. <image>Question: " + <question> + "Answer: " \\
\textbf{1-shot:} "You are a helpful medical assistant. You are being provided with images, a question about the image and an answer. Follow the example and answer the second question. <image>Question: " + <question1> + "Answer: " + <answer1> + "<|endofchunk|>" + "<image>Question: " + <question> + " Answer:" \\
\textbf{3-shot:}  "You are a helpful medical assistant. You are being provided with images, a question about the image and an answer. Follow the examples and answer the last question. <image>Question: " + <question1> + "Answer: " + <answer1> + "<|endofchunk|>" + "<image>Question: " + <question2> + "Answer: " + <answer2" + "<|endofchunk|>"+ "<image>Question: " + <question3> + "Answer: " + <answer3> + "<|endofchunk|>" + "<image>Question: " + <question> + " Answer:" \\

\subsubsection{Open-Flamingo Prompts: }
Zero-shot evaluation gives no result or generates some random tokens, which are incoherent. For \textbf{1-shot} and \textbf{3-shot}, we use the same prompts as Med-Flamingo.

\subsubsection{mPLUG=Owl Prompts: }
\textbf{Zero-shot: }"The following is a conversation between a curious human and AI assistant. The assistant gives helpful, detailed, and polite answers to the user's questions.
Human: <image>
Human: Is the facility adequate?
AI: " \\
\textbf{1-shot: }'The following is a conversation between a curious human and AI assistant. The assistant gives helpful, detailed, and polite answers to the users questions. Based on the question and answer given for the first image, give an answer for the second question based on the second image Human: <image> Human:' + <question1> +  ' AI: ' + <answer1> + 'Human: <image> Human:' + <question> + ' AI: ' \\
\textbf{3-shot: }'The following is a conversation between a curious human and AI assistant. The assistant gives helpful, detailed, and polite answers to the users questions. Based on the question and answer given for the first 3 images, give an answer for the fourth question based on the fourth image Human: <image> Human:' + <question1> +  ' AI: ' + <answer1> + 'Human: <image> Human:' + <question2> +  ' AI: ' + <answer2> + 'Human: <image> Human:' + <question3> +  ' AI: ' + <answer3> + 'Human: <image> Human:' + <question> + ' AI: '

\subsubsection{InstructBLIP Prompts}
\textbf{Zero-shot: } <image> + <question>

